# Shallow- and Deep-fake Image Manipulation Localization Using Vision Mamba and Guided Graph Neural Network

Junbin Zhang, Hamid Reza Tohidypour, Yixiao Wang and Panos Nasiopoulos

*Abstract*— Image manipulation localization is a critical research task, given that forged images may have a significant societal impact of various aspects. Such image manipulations can be produced using traditional image editing tools (known as "shallowfakes") or advanced artificial intelligence techniques ("deepfakes"). While numerous studies have focused on image manipulation localization on either shallowfake images or deepfake videos, few approaches address both cases. In this paper, we explore the feasibility of using a deep learning network to localize manipulations in both shallow- and deep-fake images, and proposed a solution for such purpose. To precisely differentiate between authentic and manipulated pixels, we leverage the Vision Mamba network to extract feature maps that clearly describe the boundaries between tampered and untouched regions. To further enhance this separation, we propose a novel Guided Graph Neural Network (G-GNN) module that amplifies the distinction between manipulated and authentic pixels. Our evaluation results show that our proposed method achieved higher inference accuracy compared to other state-of-the-art methods.

## I. INTRODUCTION

IMAGE plays a vital role in our daily life, as they convey plenty of information. As such, images that are manipulated to spread misinformation and create fraud could have a profoundly negative impact, especially when they spread online quickly and widely. The situation is even worse as the accessibility of tools for manipulating images to the public is increasing. Traditionally, manipulations were performed manually via image editing tools like Photoshop, which primarily relies on signal processing techniques. As such, this form of editing is also referred to as "shallowfakes" [1]. In recent years, the rise of deep learning and computer graphics has introduced more advanced ways for generating fake images. Among these, "deepfakes" have drawn considerable attention due to their ability to alter individuals' identities (e.g., faces) in images and videos with relative ease. With the amplification provided by social media platforms, deepfakes managed to have a huge impact, including reputational damage to celebrities, executives, and politicians. To counter the spread of such manipulated media, researchers have been actively working on developing image manipulation localization (IML) and detection techniques. Here, localization means that a solution generates masks that represent what exact regions within an image have been altered, while detection means that binary detection results for given images are provided (i.e., determining whether the images are authentic or fake).

Conventional shallowfake manipulations generally fall into three categories: splicing (copy an object from one image to another), copy-move (copy an object within an image), and inpainting (remove an object by filling it with background pixels). Each of these manipulations can distort the meaning conveyed by visual content [1]. Methods proposed for shallowfake IML ([2]-[12]) aim at finding the discrepancies between the authentic and forged areas in an image by catching certain traces left by manipulation. Besides directly using pixel domain information, these methods also utilize traces that usually reside in the high-frequency areas of images.

On the other hand, the number of approaches aiming at localizing deepfake-manipulated regions is relatively small, with the majority of existing studies concentrating on video content [13]-[15]. Similar to shallowfake solutions, these methods also attempt to find differences between authentic and manipulated areas. In addition to using traces in pixel and frequency domains, these methods rely on temporal information between video frames and/or information directly related to faces.

However, the above-mentioned IML methods are often domain-specific, targeting either shallowfakes or deepfakes. This narrow focus limits their ability to generalize across manipulation types, raising a key question: is it possible to develop a unified solution capable of localizing both?

In this paper, we present a deep learning–based approach for both detecting and localizing manipulations in shallow- and deepfake images, without requiring prior knowledge of the manipulation type. Our solution is built upon two key contributions:

- To effectively distinguish between real and manipulated regions at the pixel level, a large receptive field is essential for capturing contextual information from distant areas. To achieve this, we integrate Vision Mamba (Visual State Space Duality (VSSD) [18]) as the feature extraction backbone in

This work was supported in part by the Natural Sciences and Engineering Research Council of Canada under Grant NSERC–PG 11R12450 and in part by TELUS Corporation under Grant PG 11R10321. (*Corresponding author: Junbin Zhang*)

J. Zhang, H. R. Tohidypour, Yixiao Wang, and P. Nasiopoulos are with the University of British Columbia, Vancouver, BC, Canada. (e-mail: zjbthomas@ece.ubc.ca; htohidyp@ece.ubc.ca; yixiaow@ece.ubc.ca; panos@ece.ubc.ca).

- our framework, leveraging its ability to model long-range dependencies and enhance localization accuracy.
- To enhance the feature representation at each pixel location, we propose a novel Guided Graph Neural Network (G-GNN) that extends the original Vision GNN [19] by incorporating guided masks. Designed specifically for the IML task, these guided masks are derived from ground-truth localization labels and are used only during training to guide the network in learning more precise manipulation boundaries.

Our performance evaluation results showed that our model can achieve high inference accuracy on both shallow- and deep-fake images.

The rest of this paper is organized as follows. Section II provides an overview of related work. Section III details the proposed methodology, while Section IV presents experimental evaluation and analysis. Finally, Section V concludes the paper.

## II. RELATED WORK

### A. Image Manipulation Localization and Detection

In recent years, many works have been proposed to perform IML on images that are forged with unknown types of shallowfake techniques. Zhou et al. [2] extracted the noise distribution of an input image with Steganalysis Rich Model (SRM) filter [16] and assumed that the noise distribution between authentic and forged areas are different. They used the original RGB image and the noise distribution to localize the fake regions. Later, in ManTra-Net [3], noise distribution features are extracted by both SRM filter and BayarConv [17]. The latter proved to be more general and robust than the SRM filter [4]. The mask is generated by feeding the features into a network that detects if local features are abnormal to its neighbors. In 2020, both CR-CNN [4] and Spatial Pyramid Attention Network (SPAN) [5] are proposed that use an attention mechanism to identify discrepancies between real and fake areas. Later, Chen et al. [6] developed MVSS-Net, which considers features from the image pixels, the noise distribution extracted by BayarConv, and also identifies edges between real/fake areas for generating more accurate localization masks. Li et al. [7] also utilized edge information, and proposed a network structure that allows information in deeper layers to be injected into shallow layers for better localization. CAT-Net [8] was proposed which analyzes JPEG artifacts in manipulated images for localization. Authors of PSCC-Net [9] proposed a progressive method that generates coarse-to-fine-grained localization masks from smaller to larger scales. All the above methods are based on purely Convolutional Neural Networks (CNNs).

In recent years, Transformers have been adopted to improve the performance of networks on IML tasks. These methods include ObjectFormer [10], which first extracts features from input images and their DCT spectrum, and then utilizes a Transformer to perform localization predictions. The authors of TruFor [11] first pre-trained a network to classify types of cameras to capture and process the input images. After that, features extracted by this network are fed into a Transformer designed for semantic segmentation. Recently, Mesorch [12] is proposed, which also uses DCT to separate the images into high and low frequency components. To perform the localization tasks, the authors extracted and weighted the local and global features using a network mixed with CNNs and Transformers.

### B. Vision Mamba

For many years, solutions developed for vision related applications have been dominantly relying on CNNs and Transformers. CNNs have proved to be effective in extracting complex visual features, but they are also shown to have limited receptive fields, and, as such, they fail in capturing global information. On the other hand, Transformers are able to capture long-range visual information using self-attention mechanisms, but they suffer from quadratic computational complexity. Recently, the Mamba network [20], which is based on the idea of State Space Models (SSMs), was introduced to address the above-mentioned shortcomings of CNNs and Transformers by utilizing a global receptive field with linear computational complexity. Originally, Mamba was designed for 1D language tasks. To handle images that are in 2D space, multi-scanning mechanisms were introduced by vision Mamba models like ViM [21] and VMamba [22]. Mamba2 [23] introduces the State Space Duality (SSD) framework, enhancing model performance and efficiency. Building upon Mamba2, the Visual State Space Duality (VSSD) [18] model adapts SSD for vision tasks, eliminating the need for multi-scanning methods, and resulting in a more effective and efficient solution for image classification, detection, and segmentation tasks.

### C. Vision Graph Neural Network

Graph Neural Networks (GNNs) have become popular nowadays in handling many computer vision tasks, as graph structures help deep learning solutions to better understand complex visual relations [24]. One of the most popular GNN solutions is Vision GNN (ViG) [19]. This method directly represents images as graphs by using features of image patches as graph nodes. This solution shows great performance on multiple computer vision tasks, such as image classification [25] and video compression [26]. Graph Neural Networks (GNNs) have also been applied to image segmentation tasks. However, in most existing vision GNN based methods, the construction of the graph is not explicitly guided by semantic information. That is, all nodes - regardless of the class or semantic category they belong to - are included uniformly in the graph, without differentiation based on their labels. To the best of our knowledge, the method proposed by Hu et al. [27] is the only existing approach that performs class-wise learning in Graph Neural Networks (GNNs) for semantic segmentation. In this work, the authors introduce a class-wise dynamic graph convolution method that adaptively samples nodes with incorrect predictions. By focusing on these misclassified pixels, the model learns to emphasize regions that require correction, thereby improving segmentation accuracy.

## IV. OUR PROPOSED METHOD

To further investigate the claim made in our introduction regarding the domain-specific nature of existing IML methods, we conducted a series of empirical tests using representative shallowfake and deepfake datasets. Our goal was to evaluate how well state-of-the-art models trained on one manipulation type generalize to the other. As shown in Fig. 1, the results confirmed our hypothesis: models fine-tuned on shallowfake images struggle to accurately localize manipulations in deepfakes, and vice versa. This performance drop highlights the limited generalization capability of current IML approaches and underscores the need for a unified solution that can effectively handle both manipulation types. The following subsections outline the components of our proposed method. We first describe the overall architecture of our network, followed by the integration of Vision Mamba for robust feature extraction. We then detail the loss function employed to precisely quantify the discrepancy between the predicted and ground-truth manipulation masks.

### A. Overview of Network Design

The structure of our proposed method is shown in Fig. 2. We treat the IML task as a simplified image semantic segmentation task, with only two classes of pixels in the output mask (real or fake). As such, we decided to build our network design following the framework of UPerNet, one of the state-of-the-art image semantic segmentation networks [28]. UPerNet is a multi-task framework, which is originally designed for recognizing multiple visual concepts of a scene at once. That is to say, the network can output the category of a given scene, as well as what objects are inside the scene and the materials/textures of the objects. This multi-task framework aligns well with our objective of both detecting and localizing manipulated images. For detection, the task resembles a binary scene classification problem, where the model outputs one of two classes: real or fake. For localization, we adapt the material classification branch of the network, which leverages high-resolution, pixel-level features to produce detailed manipulation masks.

Given an input image, we first applied BayarConv [17] to extract the noise distribution of the images. This is a set of learnable high-pass filters. As shown in our previous work [29], using features extracted by BayarConv can make the network better understand the discrepancies between authentic and manipulated areas.

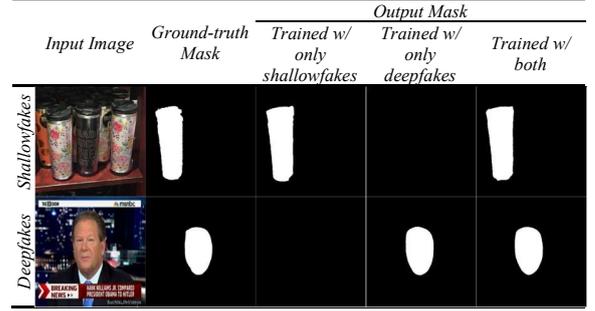

Fig. 1. Masks generated by our networks that trained on either shallow- or deep-fake images, and a mixture of both.

Our second contribution is to utilize two Vision Mamba backbone networks (VSSD) [18] to extract multi-level feature representations. We concatenate the feature maps from each level across both VSSD networks in a layer-wise manner. To effectively fuse these features, we apply a series of four simple 3×3 convolutional layers to the concatenated maps. After fusion, the highest-level information (i.e., feature maps with size 1/32 of the original image) are fed into a Pyramid Pooling Module (PPM) head [30]. As the detection task is based on the overall image-level information instead of pixel-level information, we adopted a detection head to the output of the PPM head to obtain a binary detection result (real or fake). Our detection head consists of a $3 \times 3$ convolution, an average pooling layer, and a fully connected layer.

The other levels of features are fused into the Feature Pyramid Network (FPN). At each level of FPN, we integrated our Guided Graph Neural Network (G-GNN) blocks, which are discussed in detail in Section IV. C. In the end, at the highest resolution, we attach a localization head, which outputs masks with manipulated areas highlighted. The design of the localization head is similar to the reconstruction part of many super-resolution networks [31][32], as this module requires up-sampling operations.

### B. Feature Extraction Using Vision Mamba

Separating forged from authentic regions in a manipulated image can be framed as a pixel-wise classification task, where each pixel is labeled as either real or fake, ultimately distinguishing the two types of regions. To improve the classification of real and manipulated pixels, it's important to capture not only local patterns but also global contextual relationships across the entire image. By considering how each pixel relates to all others, the network gains a more comprehensive understanding of manipulation cues. This motivates the need for our IML solution to incorporate a large receptive field, enabling it to effectively model both fine-grained and long-range dependencies. Since the recently proposed Vision Mamba architecture has demonstrated a larger Effective Receptive Field (ERF) than traditional CNN-based models, we adopt VSSD [18] - a Vision Mamba-based backbone - to extract features from both the input image and its corresponding noise map obtained using BayarConv.

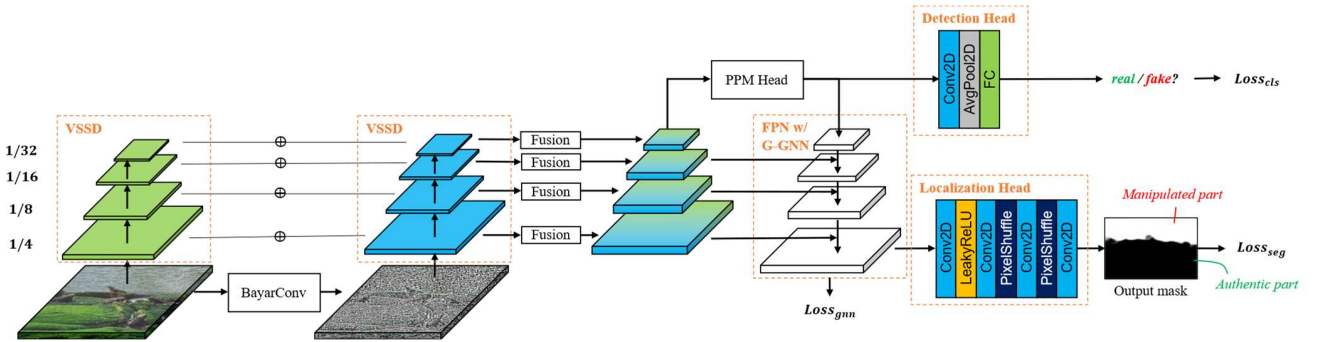

Fig. 2. Structure of our proposed solution. "⊕" means we concatenate the feature maps at each layer of the two VSSD backbones. Each concatenated feature map is then fed into a different convolutional layer for feature fusion.

Mamba [20] is based on State Space Models (SSMs), which describe the dynamics of a system. Given an input signal $x(t) \in \mathbb{R}$, we want to generate an output signal $y(t) \in \mathbb{R}$ via a hidden state $h(t) \in \mathbb{R}^N$, using the following equations:

$$h'(t) = \mathbf{A}h(t) + \mathbf{B}x(t) \\ y(t) = \mathbf{C}h(t) \tag{1}$$

where matrix $\mathbf{A} \in \mathbb{R}^{N \times N}$, $\mathbf{B} \in \mathbb{R}^{N \times 1}$ and $\mathbf{C} \in \mathbb{R}^{1 \times N}$ are learnable parameters. To adapt SSMs into deep learning system, discretization is essential. Therefore, a timescale parameter $\Delta \in \mathbb{R}$ is introduced to transform $\mathbf{A}$ and $\mathbf{B}$ into their discrete form $\overline{\mathbf{A}} = e^{\Delta \mathbf{A}}$ and $\overline{\mathbf{B}} = (\Delta \mathbf{A})^{-1}(e^{\Delta \mathbf{A}} - \mathbf{I})\Delta \mathbf{B} \approx \Delta \mathbf{B}$, with $\mathbf{I}$ as the identity matrix. Equation (1) can then be redefined as its discrete counterpart as follows:

$$h(t) = \overline{\mathbf{A}}_t h(t-1) + \overline{\mathbf{B}}_t x(t) \\ y(t) = \mathbf{C}_t h(t) \tag{2}$$

here the subscript $t$ in $\overline{\mathbf{A}}_t, \overline{\mathbf{B}}_t$, and $\mathbf{C}_t$ means that these matrices are input-dependent to $x(t)$.

Our selected backbone, VSSD, is based on Mamba2 [23], which introduces the idea of State Space Duality (SSD) so the matrix $\overline{\mathbf{A}}_t$ can be simplified as a scalar $\bar{A}_t$. [18] shows that the magnitude of $\bar{A}_t$ can be ignored so the first line of Equation (2) can be rewritten as:

$$h(t) = h(t-1) + \frac{1}{\bar{A}_t}\overline{\mathbf{B}}_t x(t) = \sum_{i=1}^{t} \frac{1}{\bar{A}_i} \overline{\mathbf{B}}_i x(t) \tag{3}$$

Until now, all the above discussions are for 1D input signals like in the case of language models. To adapt Mamba for 2D image signals, most Vision Mamba solutions split input images into $L$ multiple patches, each patch being treated as an input signal $\mathbf{X}(t)$ at time $t$. To keep the inherent structural relationships of 2D images, many Vision Mamba solutions apply multi-scanning mechanisms, i.e., the patches are still arranged as 1D arrays, but with different sequences. Since it is hard to have a perfect multi-scanning mechanism with low computational cost, VSSD proves that multi-scanning is not needed. This is because given a patch $\mathbf{X}(i)$ at time $i$, performing a forward and reverse scanning of all other patches can be calculated as:

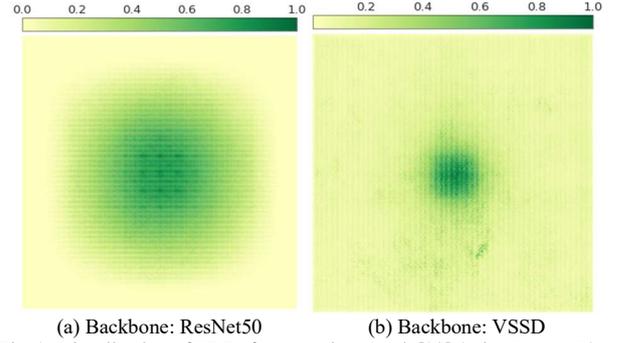

(a) Backbone: ResNet50      (b) Backbone: VSSD

Fig. 3. Visualization of ERF of our previous work [29] (using ResNet50 as the backbone) and our proposed method (using VSSD as the backbone). Note that at the boundary of (a) there are zeros, while there is no zero value in (b).

$$h(i) = \sum_{j=1}^{i} \frac{1}{\bar{A}_j}\mathbf{Z}_j + \sum_{j=-L}^{-i} \frac{1}{\bar{A}_{-j}}\mathbf{Z}_{-j} \\ = \sum_{j=1}^{L} \frac{1}{\bar{A}_j}\mathbf{Z}_j + \frac{1}{\bar{A}_i}\mathbf{Z}_i \tag{4}$$

where $\mathbf{Z}_j = \overline{\mathbf{B}}_j\mathbf{X}(j)$. VSSD considers term $\frac{1}{\bar{A}_i}\mathbf{Z}_i$ as a bias so it can be omitted. After that, all patches for different $i$ share the same hidden state, meaning that different scanning mechanisms would lead to consistent results. This makes VSSD an effective Vision Mamba method with large ERFs, so we chose it as our feature extraction backbone. In our implementation, we chose the "Micro" variant of VSSD. For the structure of VSSD, we refer readers to paper [18].

We visualize the effectiveness of our proposed method in covering a large receptive field in Fig. 3 by plotting its Effective Receptive Field (ERF), compared to the one from our previous work [29], which used ResNet50 [33] as the backbone. The displayed ERFs are averaged over 500 randomly selected images from our test set (see Section V.A). As defined in [34], ERF shows how much impact each pixel location of a set of input images has on the central pixel location. We observe that when using ResNet50 as the backbone, there are zero values at the boundary, meaning that these pixel locations have no impact on the central pixel location, leading to a small ERF only around the central area. However, when using VSSD as the backbone, there is no zero value in ERF, meaning that all pixel locations have an impact on the central pixel location, thus a bigger resultant ERF.

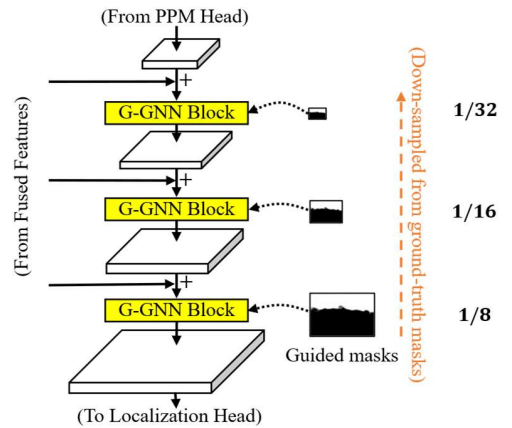

Fig. 4. Structure of our proposed G-GNN. For each level, the fused feature from the VSSD backbones and the feature from the previous level are combined together before being fed into a G-GNN block. Guided masks are obtained by down-sampling the ground-truth masks. Note that guided masks are only used during training but not testing.

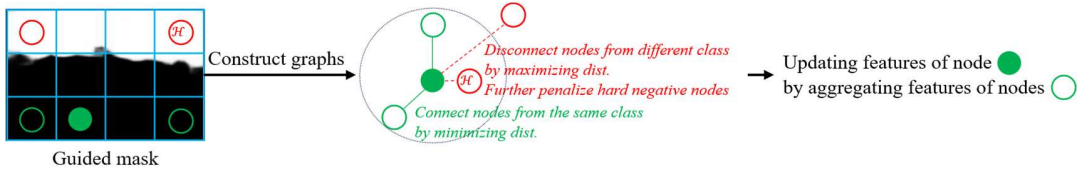

Fig. 5. Diagram of how our proposed G-GNN functions. When constructing graphs, each node is marked with its ground-truth label (real or fake) according to the guided masks provided (i.e., nodes with green color are of the same class, and are different to those with red colors). We moved nodes from the same class closer, and nodes from different class further using our proposed triplet loss, so nodes from the same class tend to be connected. Note that in the actual implementation, each node in the guided mask only represents one pixel, so there is no node that has ambiguous label.

### C. Guided Graph Neural Network (G-GNN)

The structure of our proposed G-GNN is shown in Fig. 4. In a Vision Graph Neural Network like ViG [19], which we leverage in this paper, image features are used to construct a graph. For each level $i$ of the FPN, let $\mathbf{V}_i \in \mathbb{R}^{D_i \times H_i \times W_i}$ be the fused feature obtained from the VSSD backbones, and $\mathbf{F}_i \in \mathbb{R}^{D_i \times H_i \times W_i}$ be the feature up-sampled from the last level of the FPN. We directly feed the addition $\mathbf{S}_i = \mathbf{V}_i + \mathbf{F}_i$ into the coming G-GNN block.

After reshaping $\mathbf{S}_i$ into $\mathbb{R}^{H_i W_i \times D_i}$, we followed the idea of ViG to first build a graph. When building this graph, for each node $\mathbf{s}_{ij} \in \mathbb{R}^{1 \times D_i}$, ViG connects it to nine neighbor nodes $\mathcal{C}(\mathbf{s}_{ij})$, which have closest distance to $\mathbf{s}_{ij}$ by calculating $\|\mathbf{s}_{ij} - \mathbf{s}_{ik}\|_2^2$ (where $j \neq k$). Originally, ViG finds the neighbor nodes from all nodes in $\mathbf{S}_i$, where nodes that represent different classes (i.e., one from real areas and another from fake areas) may get connected.

Therefore, we propose to use guided masks to help ViG build better graphs, and each node in the graph tends to connect to neighbor nodes from the same class to better share information. Such an idea is illustrated in Fig. 5. Only during training, we down-sample the ground-truth mask for each level of FPN using nearest neighbor interpolation to size $H_i W_i$. As such, each node in $\mathbf{S}_i$ can be labeled with the class information (real or fake) of each pixel in the down-sampled guided mask. Assuming $\mathcal{P}(\mathbf{s}_{ij})$ are all nodes in $\mathbf{S}_i$ that are of the same class as $\mathbf{s}_{ij}$ (positive nodes), and $\mathcal{N}(\mathbf{s}_{ij})$ represent different classes (negative nodes), we introduce triplet loss to minimize distance between $\mathbf{s}_{ij}$ and all $\mathcal{P}(\mathbf{s}_{ij})$, and to maximize distance between $\mathbf{s}_{ij}$ and all $\mathcal{N}(\mathbf{s}_{ij})$. In addition, we penalize negative nodes whose distance from $\mathbf{s}_{ij}$ is smaller than the furthest positive node. We label these "hard" negative nodes as $\mathcal{H}(\mathbf{s}_{ij})$. With all levels together, such triplet loss $Loss_{gnn}$ can be calculated as follows:

$$Loss_{gnn} = \sum_{i=1}^{3} \sum_{j \in \mathbf{S}_i} \left[ \left[ \sum_{k \in \mathcal{P}(\mathbf{s}_{ij})} \|\mathbf{s}_{ij} - \mathbf{s}_{ik}\|_2^2 \right] + \left[ m - \sum_{l \in \mathcal{N}(\mathbf{s}_{ij})} \|\mathbf{s}_{ij} - \mathbf{s}_{il}\|_2^2 \right] + \left[ m - \sum_{n \in \mathcal{H}(\mathbf{x}_{ij})} \|\mathbf{s}_{ij} - \mathbf{s}_{in}\|_2^2 \right] \right] \quad (5)$$

where $m$ is a fixed margin value to control the separation between positive and negative nodes. Through empirical studies, we found out that by setting $m = 10$ ensures network convergence.

After using our proposed guided solution to find a better set of neighbor nodes $\mathcal{C}'(\mathbf{s}_{ij})$ for node $\mathbf{s}_{ij}$, we utilize ViG to update the feature of this node by aggregating features from the neighbor nodes as follows:

$$\mathbf{s}'_{ij} = W_{update} \cdot g(\mathbf{s}_{ij}, \mathcal{C}'(\mathbf{s}_{ij}), W_{agg}) \quad (6)$$

where $W_{agg}$ and $W_{update}$ are learnable weights defined in ViG, and $g(\cdot)$ is a max-relative graph convolution to aggregate features of neighbor nodes:

$$g(\cdot) = conv(\mathbf{s}_{ij} \oplus \max(\{\mathbf{s}_{ik} - \mathbf{s}_{ij} | k \in \mathcal{C}'(\mathbf{s}_{ij})\}), W_{agg}) \quad (7)$$

where $\oplus$ is a concatenation operation.

In Fig. 6, we use a sample to showcase how our G-GNN blocks tend to connect a node to its neighbor nodes that come from the area of the same class. It can be seen that our G-GNN mis-connects fewer nodes than the one without guidance (0 vs 2 for real regions, and 2 vs 3 for fake regions). In addition, in G-GNN case the connections are closer to the central node, compared to the unguided GNN case.

### D. Loss Functions

Our proposed final loss function consists of three parts. On the image-level, the network should provide correct binary detection results. We used binary cross-entropy (BCE) to compute such a classification loss ($Loss_{clf}$ in Fig. 2). On the pixel-level, we used Dice loss ($Loss_{seg}$), which measures the overlap between the generated mask and a ground truth, to enhance the correctness of the generated mask. Finally, we utilize the triplet loss introduced by the G-GNN ($Loss_{gnn}$) as described by Equation (5). Our final loss is calculated as:

$$L = \alpha \cdot Loss_{clf} + \beta \cdot Loss_{seg} + \gamma \cdot Loss_{gnn} \quad (8)$$

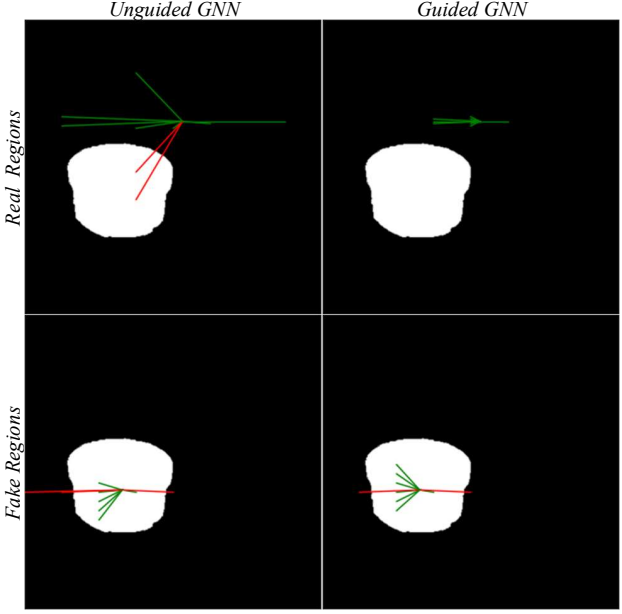

Fig. 6. Visualization of how nodes are connected in unguided GNN (left column) and our proposed guided GNN (right column). Green lines are for correctly connected pairs of nodes, while red lines are for incorrectly connected pairs.

where we set $\alpha = 0.04$, $\beta = 0.16$, and $\gamma = 0.001$. Empirical studies have shown that these coefficients yield the best localization accuracy.

## V. EXPERIMENTS AND RESULTS

### A. Experimental Setup

**Methods under comparison.** We compared our proposed solution with four state-of-the-art methods for IML: (a) CAT-Net [8], (b) ObjectFormer [10], (c) PSCC-Net [9], and (d) Mesorch [12]. We selected these methods as their training code is publicly available. We opt for methods with available training code for two reasons: to fairly compare these methods as they are originally trained using different datasets, and to evaluate their ability on detecting manipulated deepfake images.

Note that similar to our proposed method, ObjectFormer and PSCC-Net also output binary results for image manipulation detection task, while the other two (CAT-Net and Mesorch) do not.

**Dataset construction.** Our dataset consists of two parts: one for shallowfake images and one for deepfake images. The details of our training and testing sets are summarized in Table I.

*Shallowfakes.* Similar to many previous studies [6][35], we used the CASIAv2 [36] exclusively for training. This dataset includes 7,490 authentic images and 4,948 manipulated images through slicing and copy-move manipulations. We split this dataset into training, validation, and testing sets in an 8:1:1 ratio, resulting in. 5,992 real and 3,958 fake images used for training.

To evaluate the generalizability of all models, our shallowfake testing set not only includes the held-out 10% images from CASIAv2, but also images from CASIAv1 [36], Columbia [37], COVERAGE [38], and NIST16 [39]. This results in 1,832 real and 2,259 fake images, covering all three types of shallowfake manipulations (i.e., slicing, copy-move, and inpainting).

*Deepfakes.* There is no existing image dataset with ground truth masks of manipulated areas for deepfakes. To address this, we constructed a dataset using the famous FaceForensics++ [40], which provides masks for most of its deepfake videos. FaceForensics++ consists of 1,363 authentic videos from Youtube and 5,000 fake videos generated by five automated face manipulation methods: Deepfakes, Face2Face, FaceShifter, FaceSwap, and NeuralTextures.

From these videos, we extracted image frames to ensure the numbers of real and fake frames in the deepfake training and testing sets to be close to those of shallowfake ones. Specifically, we randomly selected 7 frames per real video and 2 frames per fake video generated by Deepfakes, Face2Face, FaceSwap, and NeuralTextures. We omitted 1,000 FaceShifter videos since there were no masks available for them. Due to accessibility issues, we could not download some of the videos. In total, our deepfake dataset comprises 8,449 authentic and 7,330 forged frames. These were split into training, validation,

TABLE I. OUR TRAINING AND TESTING DATASETS. FOR DEEPFAKES, NUMBERS REPRESENT THE NUMBER OF IMAGES AFTER FRAME EXTRACTION.

| | Dataset | Train | | Test | |
|---|---|---|---|---|---|
| | | #Real | #Fake | #Real | #Fake |
| Shallowfakes | CASIAv2 | 5,992 | 3,958 | 749 | 495 |
| | CASIAv1 | 0 | 0 | 800 | 920 |
| | Columbia | 0 | 0 | 183 | 180 |
| | COVERAGE | 0 | 0 | 100 | 100 |
| | NIST16 | 0 | 0 | 0 | 564 |
| | **Subtotal** | **5,992** | **3,958** | **1,832** | **2,259** |
| Deepfakes | Youtube | 5,064 | 0 | 1,688 | 0 |
| | Deepfakes | 0 | 1,000 | 0 | 600 |
| | Face2Face | 0 | 780 | 0 | 468 |
| | FaceSwap | 0 | 890 | 0 | 534 |
| | NeuralTextures | 0 | 995 | 0 | 597 |
| | **Subtotal** | **5,064** | **3,665** | **1,688** | **2,199** |
| | **Total** | **11,056** | **7,623** | **3,520** | **4,458** |

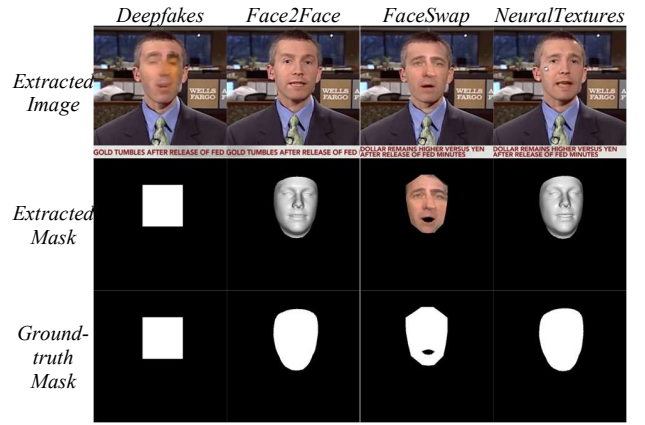

Fig. 7. Extracted and ground-truth masks in our deepfake dataset.

TABLE II. EVALUATION RESULTS OF OUR EXPERIMENTS. IN THE THIRD COLUMN, "S" MEANS TRAINING USING SHALLOWFAKE DATASET, AND "D" MEANS TRAINING USING DEEPFAKE DATASET. WE HIGHLIGHT THE BEST RESULTS IN **BOLD**, AND THE SECOND-BEST RESULTS ARE HIGHLIGHTED USING UNDERLINE.

| | Methods | Training Set | | Testing Set | | | | | | | | |
|---|---|---|---|---|---|---|---|---|---|---|---|---|
| | | | | Shallowfakes | | | Deepfakes | | | Both | | |
| | | | | Pixel-level | Image-level | | Pixel-level | Image-level | | Pixel-level | Image-level | |
| | | S | D | F1 | F1 | AUC | F1 | F1 | AUC | F1 | F1 | AUC |
| 1 | CAT-Net | √ | | 0.6272 | - | - | 0.0882 | - | - | 0.3646 | - | - |
| 2 | | | √ | 0.2706 | - | - | **0.9568** | - | - | 0.6049 | - | - |
| 3 | | √ | √ | 0.6375 | - | - | 0.8975 | - | - | <u>0.7642</u> | - | - |
| 4 | ObjectFormer | √ | | 0.5261 | **0.8028** | **0.8845** | 0.2736 | 0.2123 | 0.4020 | 0.4020 | 0.5680 | 0.6758 |
| 5 | | | √ | 0.4262 | 0.1943 | 0.4868 | 0.7851 | 0.8770 | <u>0.9604</u> | 0.6060 | 0.6158 | 0.7880 |
| 6 | | √ | √ | 0.4756 | 0.6633 | 0.7955 | 0.7046 | 0.8251 | 0.9375 | 0.5978 | 0.7471 | <u>0.8789</u> |
| 7 | PSCC-Net | √ | | 0.5746 | 0.7292 | 0.7659 | 0.0786 | 0.3050 | 0.4378 | 0.3329 | 0.5316 | 0.6060 |
| 8 | | | √ | 0.4373 | 0.0594 | 0.5084 | 0.8115 | 0.9114 | 0.9036 | 0.6196 | 0.6100 | 0.7038 |
| 9 | | √ | √ | 0.5717 | 0.6464 | 0.7184 | 0.7315 | 0.8881 | 0.8838 | 0.6496 | <u>0.7897</u> | 0.8005 |
| 10 | Mesorch | √ | | 0.6426 | - | - | 0.1771 | - | - | 0.4158 | - | - |
| 11 | | | √ | 0.4650 | - | - | 0.8386 | - | - | 0.6470 | - | - |
| 12 | | √ | √ | 0.6464 | - | - | 0.8451 | - | - | 0.7432 | - | - |
| 13 | Ours | √ | | <u>0.6717</u> | 0.7360 | 0.8657 | 0.3284 | 0.2565 | 0.2880 | 0.5044 | 0.5207 | 0.7088 |
| 14 | | | √ | 0.4425 | 0.2274 | 0.4299 | 0.9436 | **0.9719** | 0.9575 | 0.6866 | 0.6696 | 0.7995 |
| 15 | | √ | √ | **0.6830** | <u>0.7402</u> | <u>0.8810</u> | <u>0.9444</u> | <u>0.9699</u> | **0.9945** | **0.8104** | **0.8653** | **0.9320** |

TABLE III. PIXEL-LEVEL F1 RESULTS ON EACH SUBSET, WHEN METHODS ARE TRAINED ON BOTH SHALLOW- AND DEEP-FAKE TRAINING SETS. WE HIGHLIGHT THE BEST RESULTS IN **BOLD**, AND THE SECOND-BEST RESULTS ARE HIGHLIGHTED USING UNDERLINE.

| Methods | Testing Subset | | | | | | | | | |
|---|---|---|---|---|---|---|---|---|---|---|
| | Shallowfakes | | | | | Deepfakes | | | | |
| | CASIAv2 | CASIAv1 | Columbia | COVERAGE | NIST16 | Youtube | Deepfakes | Face2Face | FaceSwap | NeuralTextures |
| CAT-Net | **0.9187** | 0.6796 | 0.4933 | <u>0.4979</u> | 0.0312 | 0.8395 | **0.9345** | **0.9732** | **0.9441** | **0.9332** |
| ObjectFormer | 0.6526 | 0.4943 | 0.4037 | 0.4778 | 0.0872 | 0.6363 | 0.7652 | 0.8563 | 0.7368 | 0.6896 |
| PSCC-Net | 0.8016 | 0.6012 | 0.4294 | 0.2241 | <u>0.1898</u> | 0.6173 | 0.7718 | 0.8571 | 0.8173 | 0.8388 |
| Mesorch | 0.7737 | **0.7439** | <u>0.5935</u> | 0.4292 | 0.1792 | <u>0.8720</u> | 0.8361 | 0.8913 | 0.7681 | 0.8106 |
| Ours | <u>0.9146</u> | 0.6500 | **0.7298** | **0.6076** | **0.2696** | **0.9763** | <u>0.9223</u> | <u>0.9629</u> | <u>0.9033</u> | <u>0.8984</u> |

and testing sets with a ratio of 6:2:2 for real frames and 5:2:3 for fake frames.

The mask content varies across different deepfake manipulation methods. For instance, manipulated areas in Deepfakes are rectangle instead of face-like (we confirmed this by checking the differences between authentic videos and the corresponding manipulated videos). The other three are computer graphics-based methods, so the masks are 3D-looking (see Fig. 7, second row). For these three methods, we processed the original mask videos by discarding all colour information to generate binary ground-truth masks.

Some example masks generated for our deepfake dataset are shown in Fig. 7. For selected shallowfake datasets, readers can check the masks from their original papers [36]-[39].

**Details of training process.** We trained our solution for 100 epochs with the batch size set to 32. The learning rate for this stage was set initially to $10^{-4}$ and was decayed by a factor of 0.9 when the validation loss did not improve over 5 epochs. For optimization, we used AdamW solvers with the default momentum terms $\beta_1 = 0.9$ and $\beta_1 = 0.999$ [41]. We only applied random vertical and horizontal flipping as augmentation methods during training. Training was performed using four 32 GB NVIDIA V100 Volta GPUs on a state-of-the-art advanced research computing network [42].

We also trained the other methods we compared to on our selected training and validation sets using their default settings.

**Evaluation metrics.** The performances of networks under comparison are considered from two different angles: (1) for localization task, we reported pixel-level F1, which represents the accuracy of output masks compared to the ground-truth masks. We set the detection threshold to 0.5 for this F1 values; (2) for detection task, we reported image-level F1, which checks if the output binary results (real/fake) match the ground-truth. Similarly, the detection threshold is set to 0.5. In addition, the image-level "Area under the Receiver Operating Characteristic (ROC) Curve" (AUC) is also reported.

### B. Evaluation Results

**Comparison results.** In Table II, we report performances of the networks on all subsets from shallow- or deep-fake testing set separately, and the combination of the shallow- and deep-fake testing sets. It is clear that our proposed solution (row 15) achieves higher accuracy compared to the other state-of-the-art methods. For pixel-level accuracies, it ranks first when testing on shallowfake images, and ranks second on deepfake images. When considering both, our method is also the best, with around 5% than the second-best CAT-Net. In terms of image-level accuracy, although our method ranks just behind ObjectFormer on the shallowfake subset, it achieves the best performance across all metrics on both the deepfake subset and the overall test set.

Our experimental results also demonstrate that networks that are only trained with one of the training sets lack the ability to detect images from another set. This conforms to what we observed in Fig. 1. In addition, training with both shallow- and deep-fake datasets does not dramatically affect the accuracy on each separate set, which means that it is possible to create solutions that work for different kinds of partially manipulated images by separating pixels from real/fake regions. However, we notice that while the other state-of-the-art shallowfake IML methods show quite stable performance on shallowfake pixel-level F1 scores when using different training sets (e.g., for CAT-Net, 0.6272 in row 1 vs 0.6375 in row 3), the performance on deepfake pixel-level F1 scores may drop (e.g., also for CAT-Net, 0.9568 in row 2 vs 0.8975 in row 3). Our proposed method, on the contrary, shows more stable performance changes when different training sets were used (on shallowfakes, 0.6717 in row 13 vs 0.6830 in row 15; on deepfakes, 0.9436 in row 14 vs 0.9444 in row 15).

TABLE IV. EVALUATION RESULTS OF OUR ABLATION STUDIES. ALL METHODS ARE TRAINED ON BOTH SHALLOW- AND DEEP-FAKE TRAINING SETS. WE HIGHLIGHT THE BEST RESULTS IN **BOLD**, AND THE SECOND-BEST RESULTS ARE HIGHLIGHTED USING <u>UNDERLINE</u>.

| | Testing Set | | | | | | | | |
|---|---|---|---|---|---|---|---|---|---|
| | Shallowfakes | | | Deepfakes | | | Both | | |
| | Pixel-level | Image-level | | Pixel-level | Image-level | | Pixel-level | Image-level | |
| | F1 | F1 | AUC | F1 | F1 | AUC | F1 | F1 | AUC |
| Backbone: ResNet50 (our previous work [29]) | 0.6001 | 0.6538 | 0.8440 | 0.9097 | 0.9480 | 0.9877 | 0.7511 | 0.8170 | 0.9177 |
| Backbone: VSSD | 0.6653 | 0.7121 | <u>0.8461</u> | <u>0.9417</u> | <u>0.9672</u> | **0.9950** | 0.8000 | 0.8523 | <u>0.9260</u> |
| VSSD + GNN (no guidance) | <u>0.6735</u> | <u>0.7210</u> | 0.8303 | 0.9401 | 0.9646 | 0.9941 | <u>0.8034</u> | <u>0.8544</u> | 0.9126 |
| VSSD + G-GNN | **0.6830** | **0.7402** | **0.8810** | **0.9444** | **0.9699** | <u>0.9945</u> | **0.8104** | **0.8653** | **0.9320** |

TABLE V. PIXEL-LEVEL F1 RESULTS FOR THE ENTIRE SHALLOW- AND DEEP-FAKE TESTING SETS ON DISTORTED IMAGES. ALL METHODS ARE TRAINED ON BOTH SHALLOW- AND DEEP-FAKE TRAINING SETS. IN THE LAST COLUMN, WE AVERAGED VALUES IN EACH ROW FROM COLUMNS 2 TO 8. WE HIGHLIGHT THE BEST RESULTS IN **BOLD**, AND THE <u>SECOND-BEST RESULTS ARE HIGHLIGHTED USING UNDERLINE</u>.

| | Gaussian Noise | | | | | | | |
|---|---|---|---|---|---|---|---|---|
| **Methods** | **Standard Deviations** | | | | | | | **Avg.** |
| | **None** | **3** | **7** | **11** | **15** | **19** | **23** | |
| CAT-Net | <u>0.7642</u> | <u>0.7640</u> | <u>0.7581</u> | **0.7448** | **0.7338** | **0.7206** | **0.7141** | <u>0.7428</u> |
| ObjectFormer | 0.5978 | 0.5955 | 0.5964 | 0.5974 | 0.5927 | 0.5922 | 0.5923 | 0.5949 |
| PSCC-Net | 0.6496 | 0.6435 | 0.5394 | 0.4728 | 0.4375 | 0.4090 | 0.3976 | 0.5071 |
| Mesorch | 0.7432 | 0.7367 | 0.7210 | 0.7054 | 0.6872 | 0.6650 | 0.6492 | 0.7011 |
| Ours | **0.8104** | **0.7903** | **0.7619** | <u>0.7421</u> | <u>0.7229</u> | <u>0.7072</u> | <u>0.6931</u> | **0.7468** |

| | Gaussian Blur | | | | | | | |
|---|---|---|---|---|---|---|---|---|
| **Methods** | **Kernel Size** | | | | | | | **Avg.** |
| | **None** | **3** | **7** | **11** | **15** | **19** | **23** | |
| CAT-Net | <u>0.7642</u> | 0.6638 | 0.5864 | <u>0.5736</u> | <u>0.5622</u> | <u>0.5489</u> | **0.5380** | 0.6053 |
| ObjectFormer | 0.5978 | 0.5561 | 0.5012 | 0.4863 | 0.4777 | 0.4721 | 0.4589 | 0.5072 |
| PSCC-Net | 0.6496 | 0.4437 | 0.4173 | 0.3598 | 0.3045 | 0.2753 | 0.2546 | 0.3864 |
| Mesorch | 0.7432 | <u>0.7250</u> | **0.6908** | **0.6477** | **0.5996** | **0.5558** | 0.5282 | **0.6415** |
| Ours | **0.8104** | **0.7358** | <u>0.5938</u> | 0.5587 | 0.5453 | 0.5409 | <u>0.5374</u> | <u>0.6175</u> |

| | JPEG Compression | | | | | | | |
|---|---|---|---|---|---|---|---|---|
| **Methods** | **Quality Factors** | | | | | | | **Avg.** |
| | **None** | **100** | **90** | **80** | **70** | **60** | **50** | |
| CAT-Net | 0.7642 | 0.7663 | **0.7661** | **0.7653** | **0.7533** | **0.7268** | **0.6908** | **0.7475** |
| ObjectFormer | 0.5978 | 0.6063 | 0.5914 | 0.5704 | 0.5804 | 0.5583 | 0.5344 | 0.5770 |
| PSCC-Net | 0.6496 | 0.6134 | 0.5761 | 0.5814 | 0.5586 | 0.5069 | 0.4574 | 0.5633 |
| Mesorch | 0.7432 | 0.7351 | 0.7157 | 0.6883 | <u>0.6753</u> | <u>0.6517</u> | <u>0.6125</u> | 0.6888 |
| Ours | **0.8104** | **0.7733** | <u>0.7222</u> | <u>0.6929</u> | 0.6581 | 0.6260 | 0.5961 | 0.6970 |

In Table III we also report the pixel-level F1 values of all networks on each subset in our shallow- or deep-fake testing sets. In this case, we used the results that all networks are trained using both shallow- and deep-fake training sets. Our solution ranks first on Columbia, COVERAGE and NIST16, and Youtube subsets. Our method ranks second on the remaining six subsets, with performance only slightly below that of CAT-Net.

**Ablation studies.** In Table IV, we analyzed the effects of each important component of our proposed solution. Using our previous work [29] as a baseline, which uses ResNet50 as the feature extraction backbones, replacing the backbones to VSSD can improve both the pixel-level and image-level F1 values by around 4% when testing on the entire testing set. Moreover, introducing our G-GNN blocks can further boost both F1 values by another 1% compared to using VSSD solely, and the case of using GNN without guidance. Such improvements can be more clearly seen when evaluation involves the shallowfake testing set: around 2% improvement for pixel-level F1, 2% for image-level F1, and 4% for image-level AUC.

**Robustness to distortions.** To evaluate the robustness of the selected methods under comparison, we distorted images in our testing sets using (a) Gaussian noise (with standard deviations from 3 to 23, with a step of 4), (b) Gaussian blur (with kernel sizes from 3 to 23, with a step of 4), and (c) JPEG compression (with quality factors 100 to 50, with a step of 10). The pixel-level F1 values of networks that are trained with both shallow- and deep-fake training sets are reported in Table V. It can be seen that even though our proposed solution did not train with any distorted images as augmentation, it achieves the best average performance against Gaussian noise, and is the second best against Gaussian blur and JPEG compression. In addition, it shows high accuracy when the images are not highly distorted.

**Qualitative visualization.** For visualization purposes, in Fig. 8 we show some samples of output masks from all the networks under comparison. All these networks were trained with both shallow- and deep-fake training sets. We noticed that in most cases, our proposed solution can output masks that are visually very close to the ground-truth ones, demonstrating its good ability in handling IML task.

**Number of learnable parameters.** We show the number of learnable parameters of all networks under comparison in Table VI. While CAT-Net and Mesorch have a similar (but still lower) performance to ours, they require more learnable parameters. ObjectFormer and Mesorch have fewer learnable parameters than ours, but their accuracies are also much lower.

## VI. CONCLUSION

In this paper, we designed a novel deep learning network for detecting fake images and localizing manipulated areas in both shallowfake and deepfake images. In order to achieve this, we based our network design on UPerNet, one of the state-of-the-art image segmentation methods. We used VSSD, a Vision Mamba network, to extract features from the input images, as large ERF is important to separate pixels in real/fake areas. To further boost the accuracy of the network, we proposed G-

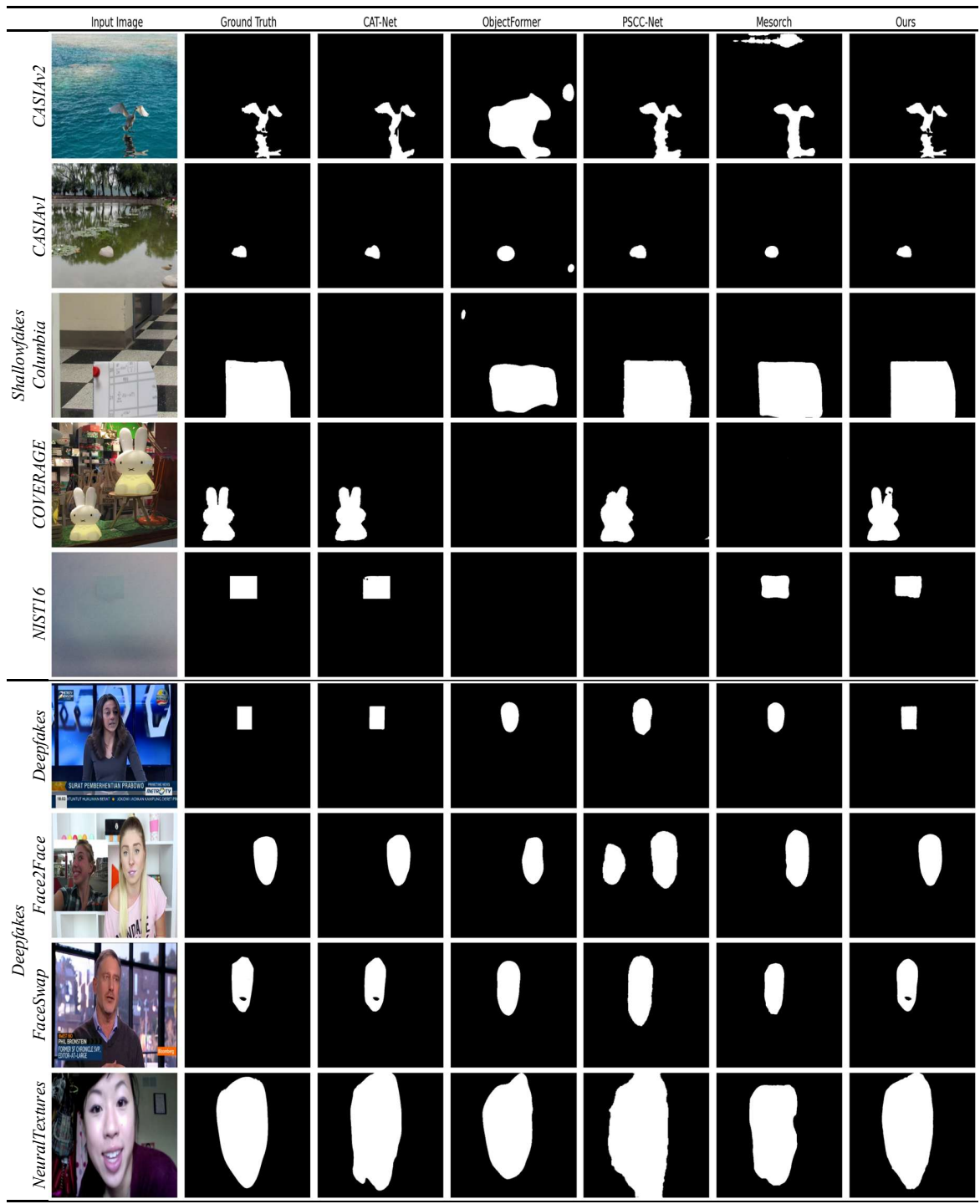

Fig. 8. Sample output masks for networks under comparison on fake image subsets, highlighting the detected modified regions.

TABLE VI. COMPARISON OF NUMBER OF LEARNABLE PARAMETERS FOR ALL NETWORKS UNDER COMPARISON (SMALLER THE BETTER).

| Methods | #Parameters (M) |
|---|---|
| CAT-Net | 114.26 |
| ObjectFormer | 34.41 |
| PSCC-Net | 3.67 |
| Mesorch | 85.75 |
| Ours | 49.50 |

GNN, which leverages ground-truth masks to guide GNN for building better graphs during training. Both our quantitative evaluation experiment results and qualitative visualization of the output masks showed that our proposed method achieved high accuracy and robustness on image manipulation localization and detection tasks.